\documentclass[letterpaper, 10 pt, conference]{ieeeconf}  

\usepackage{amsmath} 
\usepackage{amssymb} 
\usepackage{graphicx}
\usepackage{placeins} 
\usepackage{booktabs} 

\IEEEoverridecommandlockouts                              

\title{\LARGE \bf
Position-Prior-Guided Network for System Matrix Super-Resolution in Magnetic Particle Imaging
}

\author{Xuqing Geng, Lei Su, Zhongwei Bian, Zewen Sun, Jiaxuan Wen,\\ 
Jie Tian, Fellow, IEEE and Yang Du$^{*}$, Member, IEEE
\thanks{This work is supported by the National Key Research and Development Program of China 2023YFC3402804, National Natural Science Foundation of China (Grant numbers 82272111, 92159303, 62027901)}
\thanks{* is corresponding author.}
\thanks{Xuqing Geng, Lei Su, Zewen Sun, Jiaxuan Wen and Yang Du are with the CAS Key Laboratory of Molecular Imaging, Institute of Automation, Chinese Academy of Sciences, Beijing, China, and also with School of Artificial Intelligence, University of Chinese Academy of Sciences, Beijing, China
        {\tt\small yang.du@ia.ac.cn; gengxuqing2024@ia.ac.cn; sulei2023@ia.ac.cn; sunzewen2022@ia.ac.cn; wenjiaxuan2023@ia.ac.cn}%
}%
\thanks{Zhongwei Bian and Jie Tian are with the School of Engineering Medicine and the School of Biological Science and Medical Engineering, Beihang University, Beijing, China, and also with the Key Laboratory of Big Data-Based Precision Medicine (Beihang University), Ministry of Industry and Information Technology of China, Beijing, China
        {\tt\small bianzw@buaa.edu.cn; tian@ieee.org}}%
}

\begin{document}

\maketitle
\thispagestyle{empty}
\pagestyle{empty}

\begin{abstract}

Magnetic Particle Imaging (MPI) is a novel medical imaging modality. One of the established methods for MPI reconstruction is based on the System Matrix (SM). However, the calibration of the SM is often time-consuming and requires repeated measurements whenever the system parameters change. Current methodologies utilize deep learning-based super-resolution (SR) techniques to expedite SM calibration; nevertheless, these strategies do not fully exploit physical prior knowledge associated with the SM, such as symmetric positional priors. Consequently, we integrated positional priors into existing frameworks for SM calibration. Underpinned by theoretical justification, we empirically validated the efficacy of incorporating positional priors through experiments involving both 2D and 3D SM SR methods.
\newline

\indent \textit{Clinical relevance}— By reducing calibration time and enhancing resolution, the proposed approach facilitates rapid, personalized, and precise imaging across diverse individuals and conditions, thereby advancing early disease detection, vascular diagnosis, and emergency imaging.
\end{abstract}

\section{INTRODUCTION}

Magnetic Particle Imaging (MPI) is an innovative medical imaging modality capable of visualizing the spatial distribution of magnetic nanoparticles (MNPs) in vivo. This technology boasts several advantages, including unrestricted imaging depth, high sensitivity, absence of tissue signal interference, and no associated ionizing radiation risks. Such features make MPI particularly suitable for medical imaging applications, including cell imaging and angiography \cite{1,2}.

The MPI reconstruction process transforms the collected response signals from MNPs into images that represent the spatial distribution of particle concentration. Presently, there are two primary reconstruction methods: System Matrix (SM)-based reconstruction and the x-space reconstruction method \cite{3,4}. The x-space method utilizes the velocity information of the field-free point (FFP) to convert MPI data from the velocity domain to the spatial domain. Conversely, the SM method interprets the SM as a mapping between the spatial distribution of MNPs and the induced voltage received. It directly performs mathematical operations on the SM and the induced voltage to reconstruct the spatial distribution of MNPs. Compared to the x-space method, the SM method typically yields superior imaging quality \cite{15}. However, the SM measurement process is notably time-consuming. It necessitates sequential placement of a delta sample at each voxel within the imaging region, with the response signal at each point corresponding to a column of the SM. In practice, an MPI system with a limited three-dimensional field-of-view (FOV) (30 mm × 30 mm × 30 mm) is reported to take approximately 15 hours for each measurement \cite{5}. Additionally, to mitigate the impact of system noise, the SM generally requires multiple measurements for calibration, with the average value subsequently computed. Moreover, any changes in system parameters—such as variations in the type of magnetic particles or the magnetic field environment—necessitate recalibration of the SM, which significantly consumes time and resources. Consequently, it is imperative to develop methods that enable rapid acquisition of the SM for effective biomedical applications of MPI.

Research advancements have introduced several techniques to reduce SM calibration time. Among these, Compressed Sensing (CS) and deep learning-based Super-Resolution (SR) methods have emerged as prominent approaches, capable of recovering high-resolution(HR) SMs from low-resolution(LR) or down-sampled data \cite{6,7,8,9,11,12,13}. Specifically, deep learning-based approaches have demonstrated the capability to reconstruct SMs undersampled by a factor of 64, while still preserving acceptable levels of error. Despite these advancements, the quality of reconstructed SMs is still constrained by the lack of effective prior knowledge. For example, Shi et al. \cite{13} demonstrated a promising approach by incorporating receiving coil information into the reconstruction algorithm, yet there remains substantial room for improvement in SR performance by leveraging additional priors.

The SM functions as a mapping relation between the spatial distribution of MNPs and the corresponding voltage signal, exhibiting regular behavior as a function of spatial position within the FOV under ideal, noise-free conditions. In their modeling of the SM, Rahmer et al. \cite{3} demonstrated that the one-dimensional SM can be accurately characterized by the Chebyshev polynomials of the second kind. This finding suggests that the properties of the Chebyshev polynomials directly determine the spatial dependence of the SM. For ideal magnetic particles, the odd-order components in the low-dimensional SM exhibit odd symmetry, while the even-order components reflect even symmetry. Though articulating the position-dependent symmetry of the three-dimensional SM poses challenges, these observations underscore that the regularity of the SM concerning spatial information constitutes valuable prior knowledge. To harness this knowledge, we propose the incorporation of LR SM data alongside position information encoded symmetrically as priors, simultaneously feeding them into a backbone network based on RRDB-net for feature extraction. This methodology culminates in the development of the Position-Prior-Guided Network (PPGnet). Our algorithm is validated against the public OpenMPI dataset, with experimental results illustrating its robust performance across various sampling factors, thereby underscoring the potential of PPGnet for SM recovery. The primary contributions of our work are summarized as follows:

\begin{itemize}
    \item We propose PPGnet, a deep-learning network designed to accelerate SM calibration in MPI. By integrating position information with RRDBnet, the quality of the restored SM is enhanced.
\end{itemize}

\begin{itemize}
    \item The efficacy of the proposed method is evaluated on the public dataset, yielding experimental results that demonstrate its superiority over existing methods in both 2D and 3D applications.
\end{itemize}

\section{RELATED WORK}

The reconstruction of the SM is contingent upon its precise calibration. A common method for acquiring the SM involves placing a delta sample within the FOV, systematically traversing the entire volume, and recording the signal response at each point. Although this approach avoids the complexities and potential inaccuracies of physical modeling, it requires a considerable amount of time for calibration.

Current acceleration techniques focus on reconstructing high-resolution, complete SM from low-resolution or undersampled SM data. Among these techniques, interpolation-based methods are favored for their simplicity and ease of implementation. Notably, bicubic interpolation and nearest-neighbor interpolation have demonstrated commendable performance \cite{7}. However, Knopp and Weber \cite{8} introduced a novel approach by employing CS to expedite SM calibration. They initiated the process sparsifying the SM using basis transformations, such as cosine transform and discrete Fourier transform, subsequently recovering the complete SM using CS techniques. This method enabled a reduction in the number of sampling positions to 20\% a significant loss in image quality. Subsequent improvements in CS have resulted in various adaptations. For example, Weber and Knopp \cite{9} leveraged the symmetry of the SM in conjunction with CS technology to achieve a 27-fold reduction in sampling positions during calibration. Ilbey et al. \cite{10} proposed an encoded calibration scenario, which involves placing multiple MNPs samples within the FOV during each MPI scan, rather than relying on a single sample as traditional methods do. This innovation boosts the signal-to-noise ratio (SNR) and greatly enhances the effectiveness of traditional CS methods. Additionally, Gröber et al. \cite{16} utilized the low-rank properties of the SM alongside CS to facilitate accelerated calibration.

Despite the substantial progress of CS-based methods, the selection of sparse models remains largely empirical, indicating potential avenues for improvement. Drawing inspiration from the remarkable success of computer vision methodologies in image SR tasks, Baltruschat et al. \cite{6} introduced the 3D SM Recovery Network (3dSMRnet), capable of recovering 3D SM at an undersampling rate of less than 1.6\%. This approach outperformed CS methods in terms of SM recovery quality, reconstructed image fidelity, and processing time. Similarly, several deep learning frameworks for SM calibration have emerged. In 2022, Güngör et al. \cite{11} presented a Transformer-based SM SR algorithm, TranSMS, which employs dual pathways of convolution and Transformer architectures. This model achieved reduced SM recovery errors and produced higher-quality reconstructed images compared to leading CS and deep learning methods. However, data-driven approaches necessitate substantial amounts of training data, including comprehensive SM collections for training labels. To address this limitation, Yin et al. \cite{12} were pioneers in applying deep image prior technology for self-supervised learning.

To further enhance SM recovery quality, Shi et al. \cite{13} identified potential correlations between data at different frequency points of the SM and integrated prior knowledge, including receiver coil information, into their algorithms, yielding improved performance. Inspired by these findings, we propose to combine the regularity of the SM relative to positional information, integrating this positional data to enhance the performance of SM calibration models, thereby advancing current research on rapid SM calibration.

\section{METHOD}
\subsection{Symmetry of SM with Respect to Position}

Inspired by the work of Rahmer et al. \cite{3}, we investigate the symmetry inherent in modeling the SM.

\subsubsection{Signal Modeling}

In MPI, the relationship between the measured signal and particle concentration is represented by the SM. When utilizing multi-dimensional imaging sequences, signals are captured by multiple receive coils. Over the region $\Omega$, the $k$-th Fourier coefficient of the signal for the $l$-th receive channel is expressed as:

\begin{equation}
u_{l,k} = \int_{\Omega} s_{l,k}(\mathbf{r}) c(\mathbf{r}) \, d^3\mathbf{r}
\end{equation}

Here, $s_{l,k}$ denotes the $k$-th SM component for the $l$-th receive channel, and $c$ represents the particle distribution. Signal analysis is typically conducted in the frequency domain, where $k$ is referred to as the frequency index. Each SM component contains information regarding particle properties, scanner characteristics, and measurement sequences, represented by:

\begin{equation}
\hat{s}_{l,k}\left(r\right)=-{\hat{a}}_{l,k}\frac{\mu_0}{T} \int_0^T \frac{\partial}{\partial t} \left( \overline{\mathbf{m}}(\mathbf{r}, t) \cdot \mathbf{p}_l(\mathbf{r}) \right) e^{-2\pi i k t / T} \, dt
\end{equation}

In this equation, $\hat{a}_{l,k}$ is the transfer function of the receiving chain, $\overline{\mathbf{m}}(\mathbf{r}, t)$ represents the average magnetic moment generated by magnetic particles, and $\mathbf{p}_l(\mathbf{r})$ describes the coil sensitivity as a function of spatial position $\mathbf{r}$. Assuming that the superparamagnetic particles exhibit instantaneous relaxation characteristics and isotropy, the average magnetic moment can be expressed as:

\begin{equation}
\overline{\mathbf{m}}(\mathbf{r}, t) = \overline{\mathbf{m}}\left(\|\mathbf{H}(\mathbf{r}, t)\|\right) \frac{\mathbf{H}(\mathbf{r}, t)}{\|\mathbf{H}(\mathbf{r}, t)\|}
\end{equation}

To derive the symmetry of the system function, we assume that all coils are arranged orthogonally such that the coil sensitivity is represented by a unit vector and is independent of spatial position, i.e., $\mathbf{p}_l(\mathbf{r}) = \mathbf{e}_l$. Additionally, since the transfer function is independent of spatial position, it is omitted during the symmetry derivation. Under these assumptions, we have:

\begin{equation}
\tilde{m}_{l,k}(\mathbf{r}) := \frac{\hat{s}_{l,k}(\mathbf{r})}{\hat{a}_{l,k}} = -\frac{\mu_0}{T} \int_0^T \frac{\partial}{\partial t} \left( \overline{\mathbf{m}}_{l}(\mathbf{r}, t) \right) e^{-2\pi i k t / T} \, dt
\end{equation}

Since $\tilde{m}_{l,k}$ is independent of the transfer function of the receiving chain and contains only particle signal characteristics, symmetry analysis is initially performed on $\tilde{m}_{l,k}$.

\subsubsection{1D SM}

According to the derivation by Rahmer et al. \cite{3}, when $H_s(x) = Gx$, the one-dimensional SM component can be expressed as:

\begin{equation}
{\hat{s}}_{x,k}\left(r_x\right)=-\frac{2i}{T}M^\prime\left(Gr_x\right)\ast F(Gr_x)
\end{equation}

\begin{equation}
    F(Gr_x)=U_{n-1}\left(Gr_x/A\right)\sqrt{1-\left(Gr_x/A\right)^2}
\end{equation}
Thus, the one-dimensional SM component is characterized as the convolution of the derivative of the magnetization response with the second kind of Chebyshev polynomials. Since Chebyshev polynomials are even functions for odd $k$ and odd functions for even $k$, and the convolution operation propagates symmetry, we derive:

\begin{equation}
\tilde{m}_{x,k}(-r_x) = (-1)^{k+1} \tilde{m}_{x,k}(r_x)
\end{equation}

\subsubsection{2D SM}

In the two-dimensional scenario, a complete expression for the SM has yet to be derived; therefore, the symmetry of the 1D SM cannot be directly applied to the 2D case. According to the derivation by Weber et al. \cite{14}, the symmetry of the SM components arises from the symmetry of the driving field. Assuming ideal coils, isotropic superparamagnetic particles with instantaneous relaxation, and 2D driving fields where $k_x$ is even and $k_y$ is odd, we have:

\begin{equation}
{\widetilde{m}}_{x,k}\left(\left(\begin{matrix}\begin{matrix}-r_x\\r_y\\\end{matrix}\\\end{matrix}\right)\right)=\left(-1\right)^{k+1}{\widetilde{m}}_{x,k}\left(\left(\begin{matrix}\begin{matrix}r_x\\r_y\\\end{matrix}\\\end{matrix}\right)\right),k\in \mathbb{N}_0
\end{equation}

\begin{equation}
{\widetilde{m}}_{y,k}\left(\left(\begin{matrix}\begin{matrix}-r_x\\r_y\\\end{matrix}\\\end{matrix}\right)\right)=\left(-1\right)^k{\widetilde{m}}_{y,k}\left(\left(\begin{matrix}\begin{matrix}r_x\\r_y\\\end{matrix}\\\end{matrix}\right)\right), \quad k \in \mathbb{N}_0
\end{equation}

\begin{equation}
{\widetilde{m}}_{x,k}\left(\left(\begin{matrix}\begin{matrix}r_x\\{-r}_y\\\end{matrix}\\\end{matrix}\right)\right)=\left(-1\right)^{k+1}\overline{{\widetilde{m}}_{x,k}\left(\left(\begin{matrix}\begin{matrix}r_x\\r_y\\\end{matrix}\\\end{matrix}\right)\right)},\quad k \in \mathbb{N}_0
\end{equation}

\begin{equation}
{\widetilde{m}}_{y,k}\left(\left(\begin{matrix}\begin{matrix}r_x\\{-r}_y\\\end{matrix}\\\end{matrix}\right)\right)=\left(-1\right)^k\overline{{\widetilde{m}}_{y,k}\left(\left(\begin{matrix}\begin{matrix}r_x\\r_y\\\end{matrix}\\\end{matrix}\right)\right)},\quad k \in \mathbb{N}_0
\end{equation}

\subsubsection{Symmetry of SM}

\begin{equation}
{\hat{s}}_{l,k}\left(\mathbf{r}\right)={\hat{a}}_{l,k}{\widetilde{m}}_{l,k}\left(\mathbf{r}\right)
\end{equation}

From (11), the symmetry of $\hat{s}_{l,k}(\mathbf{r})$ depends on both $\hat{a}_{l,k}$ and $\tilde{m}_{l,k}(\mathbf{r})$. For symmetric equations without complex conjugation (such as (7) and (8)), symmetry can be directly utilized even if $\hat{a}_{l,k}$ is unknown. In contrast, although the symmetry of $\hat{s}_{l,k}(\mathbf{r})$ cannot be directly applied as the complex conjugate of the system function $\|\hat{s}_{l,k}(\mathbf{r})\|$, it can still be derived.

In summary, under conventional magnetic field and ideal receiving coil configurations:

\begin{itemize}
    \item For 1D SM, when $k$ is odd, $\|{\widetilde{s}}_{x,k}\|$ is even-symmetric about the origin; when $k$ is even, it is odd-symmetric about the origin.
\end{itemize}
\begin{itemize}
    \item For 2D SM, when $k$ is odd, $\|{\widetilde{s}}_{x,k}\|$ is even-symmetric about the coordinate axes, while $\|{\widetilde{s}}_{y,k}\|$ is odd-symmetric about the origin; when $k$ is even, $\|{\widetilde{s}}_{x,k}\|$ is odd-symmetric about the origin, and $\|{\widetilde{s}}_{y,k}\|$ is even-symmetric about the coordinate axes.
\end{itemize}
\begin{itemize}
    \item For 3D SM, while its symmetry cannot be explicitly derived, it is closely related to the position.
\end{itemize}

\subsection{Problem Statement}
\begin{figure*}[t]
    \centering
    \includegraphics[scale=0.3]{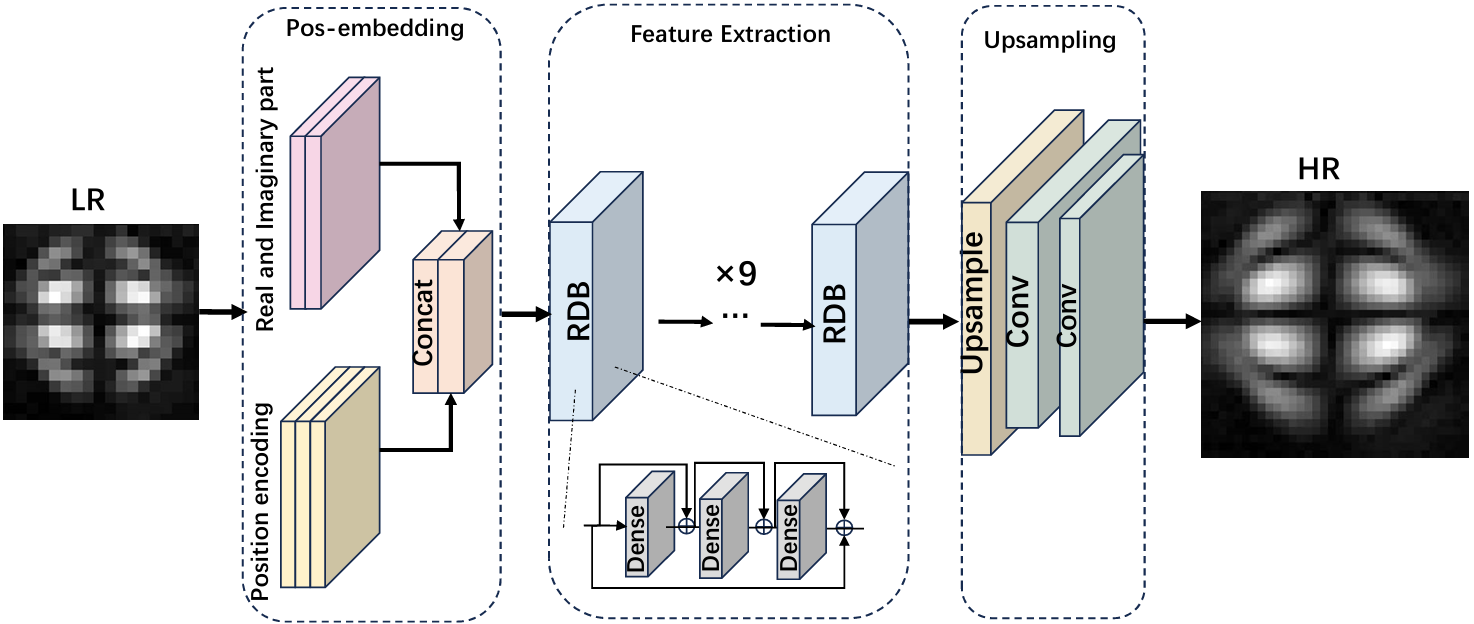}
    \caption{Architecture of the proposed PPGnet}
    \label{fig1}
\end{figure*}
Let $\mathbf{u} \in \mathbb{C}^{N_f \times 1}$ and $\mathbf{S} \in \mathbb{C}^{N_f \times N_{\text{pixel}}^H}$ represent the voltage signal and measured SM from the MPI scan, respectively, where $N_f$ and $N_{\text{pixel}}^H$ denote the total number of frequency components and the number of pixels in the HR SM. The objective of image reconstruction is to solve for the concentration of MNPs $\mathbf{c} \in \mathbb{R}^{N_{\text{pixel}}^H \times 1}$ in the equation $\mathbf{u} = \mathbf{S}\mathbf{c}$. Measuring a HR SM $\mathbf{S}$ is typically time-consuming; therefore, LR SM, denoted as $\mathbf{S}^L \in \mathbb{C}^{N_f \times N_{\text{pixel}}^L}$, is acquired to reconstruct the HR SM, $\mathbf{S}^H \in \mathbb{C}^{N_f \times N_{\text{pixel}}^H}$.

Each row of $\mathbf{S}^L$ is treated as a LR 3D image with two channels (real channel and imaginary channel). For the $n$-th row, $\mathbf{S}_n^L \in \mathbb{R}^{2 \times h \times w \times d}$, where $h \times w \times d = N_{\text{pixel}}^L$. The goal is to employ a deep learning model $f(\cdot)$ to recover $\mathbf{S}_n^H \in \mathbb{R}^{2 \times H \times W \times D}$, parameterized by $\emptyset(\theta)$ (i.e., ${\hat{s}}_n^H = f({\hat{s}}_n^L | \emptyset(\theta))$). When the single-dimension sampling factor is $r$, dimensions $H$, $W$, and $D$ are $r$ times $h$, $w$, and $d$, respectively.
\subsection{Framework}

The architecture of the proposed PPGnet is illustrated in Fig. \ref{fig1}. Three additional channels are incorporated into the input data to capture relative position information. The RRDBnet encodes the LR SM, followed by upsampling the encoded features, and ultimately employing a series of convolutional blocks to predict the HR SM components. The subsequent subsections provide detailed information about the proposed model.


\subsubsection{Position Information Embedding}
In the input layer, three additional channels $i$, $j$, and $k$ are appended to $S_i^L$, which store the relative coordinate information of SM voxels along the $x$, $y$, and $z$ axes, respectively. The $i$-coordinate channel signifies the $x$-axis information, with the first three rows in the $x$-direction filled with $0$, $1$, $2$, and so forth. Similarly, the $j$-coordinate and $k$-coordinate channels represent the $y$-axis and $z$-axis coordinate information, respectively. After populating these channels, each coordinate value is linearly scaled to fall within the range $[-1, 1]$ to complete the relative position encoding, as expressed below:

\begin{equation}
    x_n = \text{pos\_embedding}(S_n^L) \in \mathbb{R}^{5 \times h \times w \times d} 
\end{equation}

\subsubsection{Feature Extraction}
The basic architecture of SRCNN is adopted; however, to better retain and utilize low-level features, the basic convolutional blocks are replaced with Residual Dense Blocks (RDBs). Furthermore, all 2D convolutions are substituted with 3D convolutions, extending the network from 2D image processing to 3D SM processing. Each RDB consists of three dense blocks and four residual connections, where each dense block includes five convolution layers and their corresponding activation functions. Following feature extraction, the SM features $z_n$ are obtained as follows:

\begin{equation}
    z_n = \text{RRDBnet}(x_n) \in \mathbb{R}^{C' \times h \times w \times d} 
\end{equation}

\subsubsection{Upsampling}
First, HR feature maps $S_n^H$ are derived through an upsampling module, followed by a convolution module to obtain the HR SM $\hat{S}_n^H \in \mathbb{R}^{C \times H \times W \times D}$ from the feature maps. The upsampling module employs linear interpolation, as illustrated below:
\begin{equation}
    S_n^H = \text{Upsampling}(z_n) \in \mathbb{R}^{C' \times H \times W \times D}  
\end{equation}
\begin{equation}
    \hat{S}_n^H = \text{Conv3D}(S_n^H) \in \mathbb{R}^{C \times H \times W \times D}  
\end{equation}

\section{Experiment}
\subsection{Experimental Setup}

\subsubsection{Dataset}

This experiment employs the OpenMPI dataset, the inaugural open-source MPI dataset. It comprises SM calibration and phantom measurement data collected using various MNPs. For the SM recovery experiment, we selected the SM calibration data with serial number 7 from Synomag-D, referred to as $S_{HR}^{Syno}$, to construct the training set. Additionally, we utilized the SM calibration data with serial number 6 from Perimag, denoted as $S_{HR}^{Peri}$, as the test set to evaluate the model's performance. This configuration enables an assessment of the algorithm's generalization capability across different types of MNPs. Both $S_{HR}^{Syno}$ and $S_{HR}^{Peri}$ were obtained under conditions using a 4$\mu$L concentration of 100mmol/L delta samples with a grid size of 37×37×37, resulting in sizes of 37×37×37×$K$. To eliminate base frequency and mitigate the effects of acquisition noise, only SM rows with frequencies exceeding 8 kHz and a SNR greater than 3 were retained. After processing, $S_{HR}^{Syno}$ and $S_{HR}^{Peri}$ contained $K=3929$ and $K=3175$ frequency components, respectively. For the phantom reconstruction experiment, both resolution and shape phantoms from the OpenMPI dataset were selected.

\subsubsection{Implementation Details}

To generate LR and HR SM pairs for training, zero-padding was applied to $S_{HR}^{Syno}$ and $S_{HR}^{Peri}$,  adding one row at the beginning and two rows at the end to achieve a size of 40×40×40×$K$. Subsequently, 2× and 4× LR SMs of sizes 20×20×20×$K$ and 10×10×10×$K$ were obtained through equidistant sampling in each dimension. For 2D experiments, the z-channel data was flattened into 37×37×37×$K$, followed by zero-padding and downsampling operations similar to those in 3D to obtain 2D LR and HR SM pairs. During training, $K$ of $S_{HR}^{Syno}$ was divided into 90\% training data and 10\% validation data.

Parameters were optimized separately for each downsampling factor (2×, 4×) to maximize validation performance in terms of normalized root mean square error (NRMSE). During network training, random 90° rotations and flips were employed as data augmentation techniques without regularization. The model was implemented in PyTorch and trained on an NVIDIA GeForce RTX 4090 GPU, with the final model selected based on the best validation NRMSE. For image reconstruction using calibrated SMs, the Kaczmarz regularization algorithm was employed with $\lambda = 0.75$ and 3 iterations.

\subsubsection{Comparison Methods}

The proposed method was compared with existing techniques for accelerating SM calibration, including methods for both 2D and 3D SM. 3dSMRnet represents the first deep learning-based approach for 3D SM calibration, leveraging residual dense blocks to extract features from LR SM and subsequently upsampling feature maps to reconstruct HR SM using 3D convolutions. TranSMS is a sophisticated 2D SM calibration model characterized by a dual-branch architecture comprising convolutional and Transformer branches, where the Transformer branch incorporates Transformer modules with convolution-based block embedding methods.

\subsubsection{Evaluation Metrics}

For every experiment, both the baseline models and the proposed model needed an identical number of calibration measurements. NRMSE was employed as the evaluation metric for SM calibration, defined as follows:
\begin{equation}
    \text{NRMSE}\left(\hat{s}_i^H\right)=\frac{\|\hat{s}_i^H-s_i^H\|_F}{\max{\left(|s_i^H|\right)}-\min{\left(|s_i^H|\right)}}
\end{equation}
where $\|\cdot\|_F$ denotes the Frobenius norm, and $|\cdot|$ represents the complex modulus, with values converted to a complex format for evaluation. To assess the quality of reconstructed images, we computed the Peak Signal-to-Noise Ratio (PSNR), Structural Similarity Index (SSIM), and NRMSE.

\subsection{System Matrix Recovery}
\subsubsection{3D Experiments}

Table~\ref{tab1} presents the results of 2D and 3D SM calibration. For the 3D results, the proposed method consistently outperforms the similarly structured 3dSMRnet at each sampling rate. Specifically, with a downsampling factor of 2, the proposed PPGnet achieves an approximate 33\% improvement in NRMSE compared to 3dSMRnet. At a downsampling factor of 4, PPGnet continues to show a notable enhancement, reducing NRMSE by about 13\% relative to 3dSMRnet. These findings indicate that the proposed method demonstrates superior performance under a comparable architectural framework.

\begin{figure}[t]
    \centering
    \includegraphics[scale=0.28]{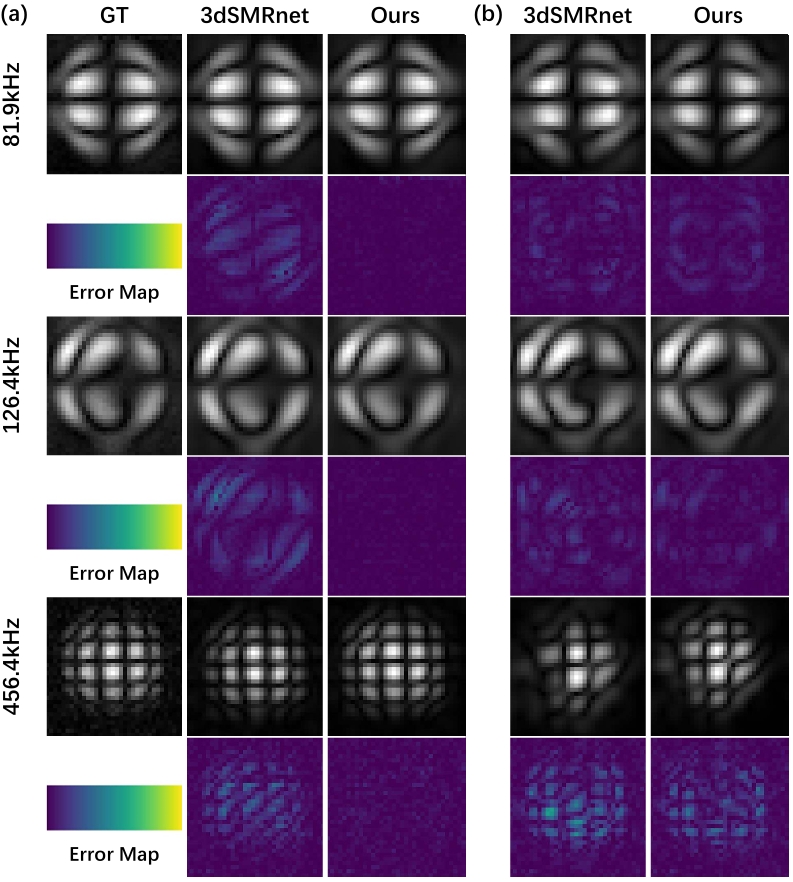} 
    \caption{The visual representation of three reconstructed 3D SM rows (central slice) for downsampling ratios of 2 (a) and 4 (b), respectively.}
    \label{fig2}
\end{figure}

\renewcommand{\arraystretch}{1.5}
\setlength{\tabcolsep}{15pt}
\begin{table}[t]
    \centering
    \caption{SM Calibration Results}
    \label{tab1}
    \begin{tabular}{lcc}
        \toprule
        \textbf{Ratio} & \multicolumn{1}{c}{$2\times$} & \multicolumn{1}{c}{$4\times$} \\
        \textbf{Method} & $\overline{\text{NRMSE}} \downarrow$ & $\overline{\text{NRMSE}} \downarrow$ \\
        \midrule
        
        3dSMRnet & 5.14\% & 5.95\% \\
        3d-PPGnet & \textbf{3.44\%} & \textbf{5.17\%} \\
        TranSMS & 4.08\% & 5.20\% \\
        2d-PPGnet & \textbf{3.14\%} & \textbf{4.71\%} \\
        \bottomrule
    \end{tabular}
\end{table}
Fig.\ref{fig2} shows the central slice of the reconstructed 3D SM data for visual assessment. Three central slices from different frequencies of the SM are displayed, accompanied by the corresponding error maps. Overall, the recovery performance of the proposed method surpasses that of 3dSMRnet. Notably, as depicted in the fifth row of Fig.\ref{fig2}(b), severe damage resulting from excessive undersampling causes significant distortions in the SM shapes recovered by 3dSMRnet. In contrast, the proposed method effectively incorporates positional priors, enabling the recovery of severely damaged structures.

\subsubsection{2D Experiments}

\begin{figure}[tb]
    \centering
    \includegraphics[scale=0.28]{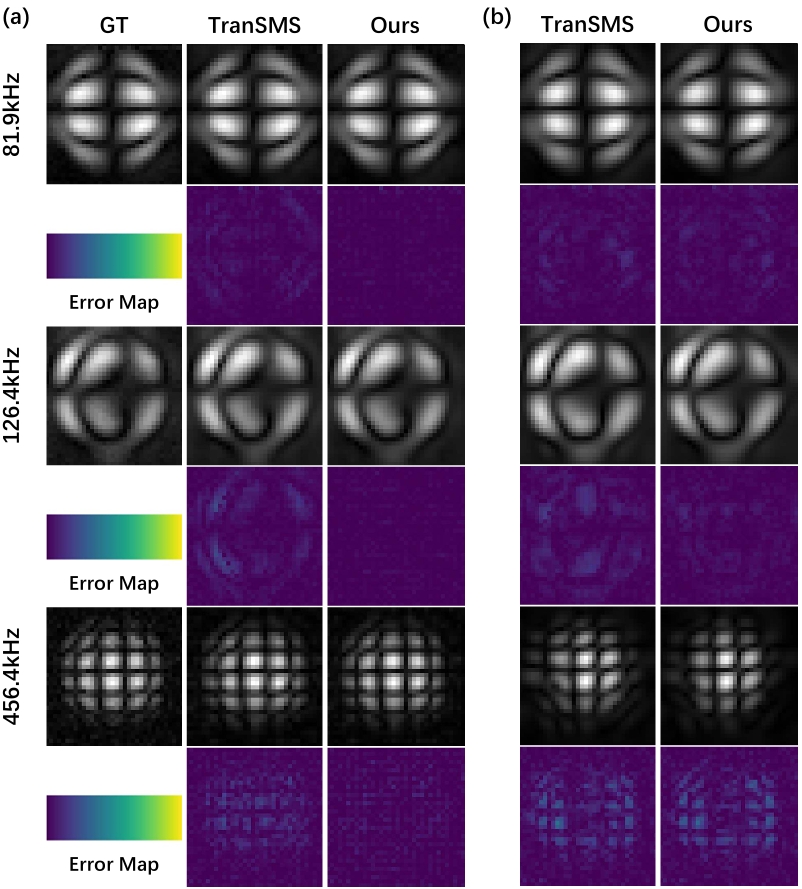} 
    \caption{The visual representation of three reconstructed 2D SM rows (central slice) for downsampling ratios of 2 (a) and 4 (b), respectively.}
    \label{fig3}
\end{figure}

To benchmark against the leading model TranSMS for 2D SM calibration, we employed the developed PPGnet to process the 2D data. The dataset used was identical to that of the 3D experiment, except that the 3D dataset was flattened into a 2D format for this analysis. The SM calibration results are presented in Table~\ref{tab1}. In the SM recovery experiment, PPGnet outperformed TranSMS at both sampling rates. At a 2× sampling rate, the ${\text{NRMSE}}$ achieved an optimal value of 3.14\%. Consistent with the 3D experiments, Fig.~\ref{fig3} shows the visual results of SM recovery, confirming the superiority of the proposed method over TranSMS in the 2D context.

\subsection{Image Reconstruction}

Tables~\ref{tab3} and \ref{tab4} present the image reconstruction outcomes of calibrated SM on the OpenMPI dataset for two different phantoms. According to Table~\ref{tab4}, the reconstruction results obtained using different methods align with the SM recovery results. Our proposed method demonstrates significant advancements compared to both 3dSMRnet and TranSMS.

\begin{figure}[t]
    \centering
    \includegraphics[scale=0.3]{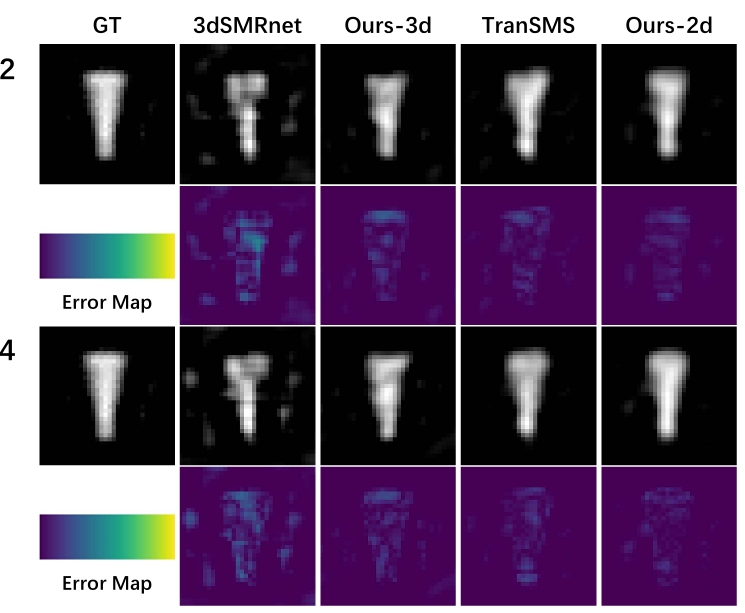} 
    \caption{The image reconstruction outcome for the shape phantom. The first row displays the reconstructed image, while the second row shows the corresponding 3D error map averaged along the Z-axis. The numbers “2” and “4” represent the downsampling ratios. The GT image is reconstructed using the measured full-size SM.}
    \label{fig4}
\end{figure}

Fig.~\ref{fig4} illustrates the image reconstruction outcome for the shape phantom, accompanied by the corresponding error maps. The darker error maps of PPGnet validate its more reliable reconstruction compared to the other methods.

\renewcommand{\arraystretch}{1.5}
\setlength{\tabcolsep}{2pt}
\begin{table}[t]
    \centering
    \caption{Reconstruction Results of Calibrated  SM on Resolution Phantom}
    \label{tab3}
    \begin{tabular}{lcccccc}
        \toprule
        \textbf{Ratio} & \multicolumn{3}{c}{$2\times$} & \multicolumn{3}{c}{$4\times$} \\
        \cmidrule(lr){2-4} \cmidrule(lr){5-7}
        \textbf{Method} & $\overline{\text{NRMSE}} \downarrow$ & PSNR $\uparrow$ & SSIM $\uparrow$ & $\overline{\text{NRMSE}} \downarrow$ & PSNR $\uparrow$ & SSIM $\uparrow$ \\
        \midrule
        3dSMRnet & 5.59\% & 25.04 & 0.4857 & 5.65\% & 24.96 & 0.4834 \\
        3d-PPGnet &\textbf{4.62\%} & \textbf{26.71} & \textbf{0.5303} & \textbf{4.65\%} & \textbf{26.65} & \textbf{0.5260} \\
        TranSMS & 3.76\% & 28.49 & 0.6278 & 3.93\% & 28.10 & 0.6242 \\
        2d-PPGnet & \textbf{3.24\%} & \textbf{29.80} & \textbf{0.6581} & \textbf{3.26\%} & \textbf{29.72} & \textbf{0.6572} \\
        \bottomrule
    \end{tabular}
\end{table}

\renewcommand{\arraystretch}{1.5}
\setlength{\tabcolsep}{2pt}
\begin{table}[t]
    \centering
    \caption{Reconstruction Results of Calibrated SM on Shape Phantom}
    \label{tab4}
    \begin{tabular}{lcccccc}
        \toprule
        \textbf{Ratio} & \multicolumn{3}{c}{$2\times$} & \multicolumn{3}{c}{$4\times$} \\
        \cmidrule(lr){2-4} \cmidrule(lr){5-7}
        \textbf{Method} & $\overline{\text{NRMSE}} \downarrow$ & PSNR $\uparrow$ & SSIM $\uparrow$ & $\overline{\text{NRMSE}} \downarrow$ & PSNR $\uparrow$ & SSIM $\uparrow$ \\
        \midrule
        3dSMRnet & 5.19\% & 25.70 & 0.5574 & 5.99\% & 24.45 & 0.5251 \\
        3d-PPGnet & \textbf{4.19\%} & \textbf{27.55} & \textbf{0.6505} & \textbf{4.27\% }& \textbf{27.39} & \textbf{0.6327} \\
        TranSMS & 2.66\% & 31.51 & 0.7861 & 2.76\% & 31.19 & 0.7801 \\
        2d-PPGnet & \textbf{2.13\%} & \textbf{33.45} & \textbf{0.8253} & \textbf{2.32\%} & \textbf{32.70} & \textbf{0.8163} \\
        \bottomrule
    \end{tabular}
\end{table}

\subsection{Ablation Study}

\begin{figure}[!t]
    \centering
    \includegraphics[scale=0.28]{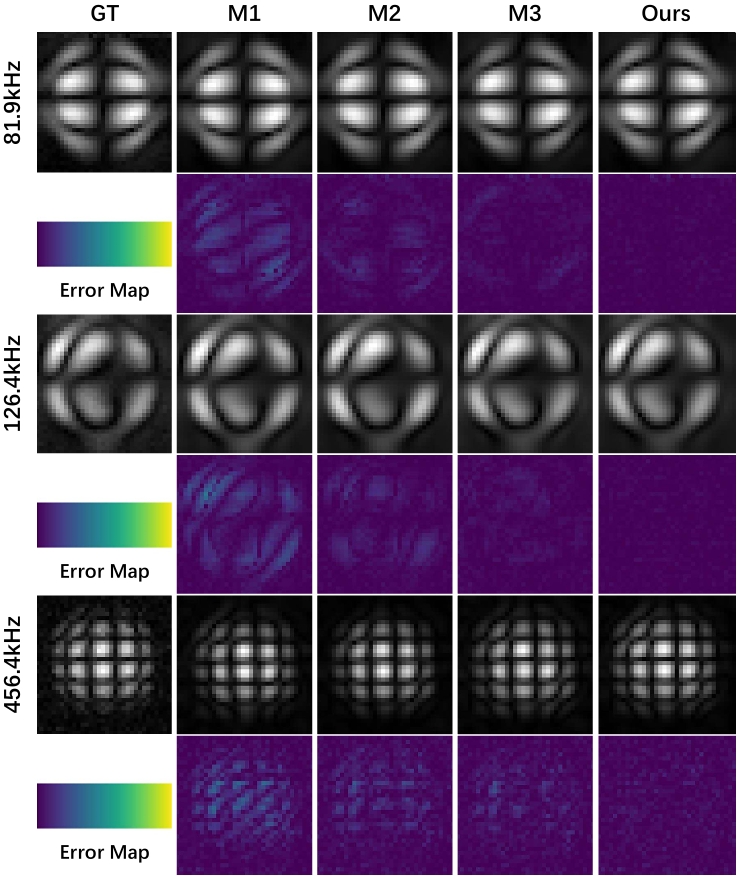}
    \caption{Visualization results of SM calibration from ablation study (2$\times$ sampling).}
    \label{fig6}
\end{figure}

In this section, we conduct ablation studies to investigate the impact of key modules, specifically the presence or absence of position encoding, the type of position encoding used, and the upsampling methods applied. The model configurations are as follows: no position encoding (M1), normalized position encoding (M2), symmetric position encoding (M3), all utilizing nearest-neighbor interpolation for upsampling. Finally, PPGnet employs symmetric position encoding in conjunction with linear interpolation for upsampling. The SM calibration and image reconstruction results are presented in Tables~\ref{tab7} and \ref{tab9}. Additionally, Fig.~\ref{fig6} depicts the central slice of the reconstructed SM data for visual assessment. By comparing experimental outcomes and visualizations, we show how these modules effectively improve SM recovery and the quality of image reconstruction.

\setlength{\tabcolsep}{8pt}
\begin{table}[b]
    \centering
    \caption{Ablation Study Results on SM Calibration}
    \label{tab7}
    \begin{tabular}{lcc}
        \toprule
        Ratio &{$2\times$} & {$4\times$} \\
        Method & $\overline{\text{NRMSE}}\downarrow$ & $\overline{\text{NRMSE}}\downarrow$ \\
        \midrule
        base(M1) & 5.14\% & 5.95\% \\
        base+norm(M2) & 4.29\% & 5.69\% \\
        base+sym(M3) & 4.17\% & 5.27\% \\
        \textbf{base+sym+tri(PPGnet)} & \textbf{3.44\% }& \textbf{5.17\%} \\
        \bottomrule
    \end{tabular}
\end{table}

\setlength{\tabcolsep}{1pt}
\begin{table}[tb]
    \centering
    \caption{Ablation Study Results on Image Reconstruction for Shape Phantom}
    \label{tab9}
    \begin{tabular}{lcccccc}
        \toprule
        Ratio & \multicolumn{3}{c}{2$\times$} & \multicolumn{3}{c}{4$\times$} \\
        \cmidrule(lr){2-4} \cmidrule(lr){5-7}
        Method & $\overline{\text{NRMSE}}\downarrow$ & PSNR$\uparrow$ & SSIM$\uparrow$ & $\overline{\text{NRMSE}}\downarrow$ & PSNR$\uparrow$ & SSIM$\uparrow$ \\
        \midrule
        M1 & 5.19\% & 25.70 & 0.5574 & 5.99\% & 24.45 & 0.5251 \\
        M2 & 6.61\% & 23.59 & 0.5178 & 7.22\% & 22.83 & 0.4920 \\
        M3 & 6.28\% & 24.05 & 0.5281 & 6.30\% & 24.01 & 0.4828 \\
        PPGnet & \textbf{4.19\%} & \textbf{27.55} & \textbf{0.6505} & \textbf{4.27\%} & \textbf{27.39} & \textbf{0.6327} \\
        \bottomrule
    \end{tabular}
\end{table}

\section{Discussion}

We propose a SR SM recovery method that integrates positional information to enhance recovery quality and reduce calibration time. Our results show that combining symmetric position encoding with linear interpolation significantly improves recovery performance, highlighting the method’s potential to advance SM calibration and reconstruction.

While the system symmetry we derived assumes ideal conditions, real-world MPI systems are subject to noise, calibration errors, and parameter variations. These factors may affect performance, indicating the need for further research into the impact of noise and other practical challenges on the method’s robustness.
\section{Conclusion}

We present an innovative SR SM recovery method that significantly enhances reconstruction quality by incorporating positional prior information. We anticipate that this work will inspire further research and exploration aimed at improving MPI performance, thereby promoting its widespread application in biomedical imaging and related fields.

\addtolength{\textheight}{-12cm}   




\section*{ACKNOWLEDGMENT}

The authors would like to thank the instrumental and technical support of Multimodal Biomedical Imaging Experimental Platform, Institute of Automation, Chinese Academy of Sciences.

\bibliographystyle{IEEEtran}
\bibliography{ref}

@article{1,
  title={Cellular uptake of magnetic nanoparticles imaged and quantified by magnetic particle imaging},
  author={Paysen, Hendrik and Loewa, Norbert and Stach, Anke and Wells, James and Kosch, Olaf and Twamley, Shailey and Makowski, Marcus R and Schaeffter, Tobias and Ludwig, Antje and Wiekhorst, Frank},
  journal={Scientific reports},
  volume={10},
  number={1},
  pages={1922},
  year={2020},
  publisher={Nature Publishing Group UK London}
}

@article{2,
  title={Magnetic particle imaging: a novel in vivo imaging platform for cancer detection},
  author={Yu, Elaine Y and Bishop, Mindy and Zheng, Bo and Ferguson, R Matthew and Khandhar, Amit P and Kemp, Scott J and Krishnan, Kannan M and Goodwill, Patrick W and Conolly, Steven M},
  journal={Nano letters},
  volume={17},
  number={3},
  pages={1648--1654},
  year={2017},
  publisher={ACS Publications}
}

@article{3,
  title={Signal encoding in magnetic particle imaging: properties of the system function},
  author={Rahmer, J{\"u}rgen and Weizenecker, J{\"u}rgen and Gleich, Bernhard and Borgert, J{\"o}rn},
  journal={BMC medical imaging},
  volume={9},
  pages={1--21},
  year={2009},
  publisher={Springer}
}

@article{4,
  title={The X-space formulation of the magnetic particle imaging process: 1-D signal, resolution, bandwidth, SNR, SAR, and magnetostimulation},
  author={Goodwill, Patrick W and Conolly, Steven M},
  journal={IEEE transactions on medical imaging},
  volume={29},
  number={11},
  pages={1851--1859},
  year={2010},
  publisher={IEEE}
}

@article{5,
  title={Magnetic particle imaging: from proof of principle to preclinical applications},
  author={Knopp, Tobias and Gdaniec, Nadine and M{\"o}ddel, Martin},
  journal={Physics in Medicine \& Biology},
  volume={62},
  number={14},
  pages={R124},
  year={2017},
  publisher={IOP Publishing}
}

@inproceedings{6,
  title={3d-SMRnet: Achieving a new quality of MPI system matrix recovery by deep learning},
  author={Baltruschat, Ivo M and Szwargulski, Patryk and Griese, Florian and Grosser, Mirco and Werner, Rene and Knopp, Tobias},
  booktitle={Medical Image Computing and Computer Assisted Intervention--MICCAI 2020: 23rd International Conference, Lima, Peru, October 4--8, 2020, Proceedings, Part II 23},
  pages={74--82},
  year={2020},
  organization={Springer}
}

@article{7,
  title={Super-resolving reconstruction technique for MPI},
  author={Gungor, Alper and Top, Can Bar{\i}{\c{s}}},
  journal={International Journal on Magnetic Particle Imaging IJMPI},
  volume={6},
  number={2 Suppl 1},
  year={2020}
}

@article{8,
  title={Sparse reconstruction of the magnetic particle imaging system matrix},
  author={Knopp, Tobias and Weber, Alexander},
  journal={IEEE transactions on medical imaging},
  volume={32},
  number={8},
  pages={1473--1480},
  year={2013},
  publisher={IEEE}
}

@article{9,
  title={Reconstruction of the magnetic particle imaging system matrix using symmetries and compressed sensing},
  author={Weber, Alexander and Knopp, Tobias},
  journal={Advances in Mathematical Physics},
  volume={2015},
  number={1},
  pages={460496},
  year={2015},
  publisher={Wiley Online Library}
}

@article{10,
  title={Fast system calibration with coded calibration scenes for magnetic particle imaging},
  author={Ilbey, Serhat and Top, Can Bar{\i}{\c{s}} and G{\"u}ng{\"o}r, Alper and {\c{C}}ukur, Tolga and Saritas, Emine Ulku and G{\"u}ven, H Emre},
  journal={IEEE transactions on medical imaging},
  volume={38},
  number={9},
  pages={2070--2080},
  year={2019},
  publisher={IEEE}
}

@article{11,
  title={TranSMS: Transformers for super-resolution calibration in magnetic particle imaging},
  author={G{\"u}ng{\"o}r, Alper and Askin, Baris and Soydan, Damla Alptekin and Saritas, Emine Ulku and Top, Can Bar{\i}{\c{s}} and {\c{C}}ukur, Tolga},
  journal={IEEE Transactions on Medical Imaging},
  volume={41},
  number={12},
  pages={3562--3574},
  year={2022},
  publisher={IEEE}
}

@article{12,
  title={System matrix recovery based on deep image prior in magnetic particle imaging},
  author={Yin, Lin and Guo, Hongbo and Zhang, Peng and Li, Yimeng and Hui, Hui and Du, Yang and Tian, Jie},
  journal={Physics in Medicine \& Biology},
  volume={68},
  number={3},
  pages={035006},
  year={2023},
  publisher={IOP Publishing}
}

@article{13,
  title={Progressive pretraining network for 3D system matrix calibration in magnetic particle imaging},
  author={Shi, Gen and Yin, Lin and An, Yu and Li, Guanghui and Zhang, Liwen and Bian, Zhongwei and Chen, Ziwei and Zhang, Haoran and Hui, Hui and Tian, Jie},
  journal={IEEE Transactions on Medical Imaging},
  year={2023},
  publisher={IEEE}
}

@article{14,
  title={Symmetries of the 2D magnetic particle imaging system matrix},
  author={Weber, Alexander and Knopp, Tobias},
  journal={Physics in Medicine \& Biology},
  volume={60},
  number={10},
  pages={4033},
  year={2015},
  publisher={IOP Publishing}
}

@article{15,
  title={Recent developments of the reconstruction in magnetic particle imaging},
  author={Yin, Lin and Li, Wei and Du, Yang and Wang, Kun and Liu, Zhenyu and Hui, Hui and Tian, Jie},
  journal={Visual computing for industry, biomedicine, and art},
  volume={5},
  number={1},
  pages={24},
  year={2022},
  publisher={Springer}
}

@article{16,
  title={Enhanced compressed sensing recovery of multi-patch system matrices in MPI},
  author={Grosser, Mirco and Boberg, Marija and Bahe, Marleen and Knopp, Tobias},
  journal={International Journal on Magnetic Particle Imaging IJMPI},
  volume={6},
  number={2 Suppl 1},
  year={2020}
}

\end{document}